\pdfoutput=1

\documentclass[11pt]{article}

\usepackage[]{acl}

\usepackage{times}
\usepackage{latexsym}

\usepackage[T1]{fontenc}

\usepackage[utf8]{inputenc}

\usepackage{microtype}

\usepackage{inconsolata}

\usepackage{graphicx}               
\usepackage{tabularx}               
\usepackage{booktabs}
\usepackage{amsmath}
\usepackage{stmaryrd}
\usepackage{xcolor}
\usepackage{soul}

\newcolumntype{C}{>{\centering\arraybackslash}X}
\usepackage{multirow}               
\usepackage{diagbox}                
\usepackage{hhline}                 
\usepackage{color}                  
\usepackage{amsmath}                
\usepackage{amssymb}                
\usepackage{mathtools}              
\usepackage{enumitem}               

\usepackage{subfigure}
\usepackage{booktabs}
\usepackage{extarrows}
\usepackage{makecell}

\usepackage{soul}

\definecolor{RoseQuartzBg}{HTML}{F7CAC9}
\definecolor{RoseQuartz}{HTML}{F5A798}
\definecolor{Serenity}{HTML}{92A8D1}
\definecolor{OrangeRed}{rgb}{1.0, 0.27, 0.0}
\definecolor{Red}{rgb}{1.0, 0.0, 0.0}
\definecolor{Turquoise}{HTML}{0F4C81}
\definecolor{RoyalBlue}{cmyk}{1, 0.50, 0, 0}
\definecolor{Green}{rgb}{0.0, 1.0, 0.0}
\definecolor{Orchid}{rgb}{0.85, 0.44, 0.84}
\definecolor{Orange}{rgb}{1.0, 0.5, 0.0}
\definecolor{darkpink}{rgb}{0.91, 0.33, 0.5}
\usepackage{xparse}
\NewDocumentCommand{\lifu}{ mO{} }{\textcolor{Red}{\textsuperscript{\textit{Lifu}}\textsf{\textbf{\small[#1]}}}}
\NewDocumentCommand{\barry}{ mO{} }{\textcolor{blue}{\textsuperscript{\textit{Barry}}\textsf{\textbf{\small[#1]}}}}
\NewDocumentCommand{\hugo}{ mO{} }{\textcolor{Serenity}{\textsuperscript{\textit{Hugo}}\textsf{\textbf{\small[#1]}}}}
\NewDocumentCommand{\qf}{ mO{} }{\textcolor{RoseQuartzBg}{\textsuperscript{\textit{Qifan}}\textsf{\textbf{\small[#1]}}}}
\NewDocumentCommand{\sijia}{ mO{} }{\textcolor{RoyalBlue}{\textsuperscript{\textit{sijia}}\textsf{\textbf{\small[#1]}}}}
\NewDocumentCommand{\minqian}{ mO{} }{\textcolor{teal}{\textsuperscript{\textit{Minqian}}\textsf{\textbf{\small[#1]}}}}
\NewDocumentCommand{\zhiyang}{ mO{} }{\textcolor{OrangeRed}{\textsuperscript{\textit{Zhiyang}}\textsf{\textbf{\small[#1]}}}}

\usepackage{amssymb}
\usepackage{pifont}
\newcommand{\cmark}{\ding{51}}%
\newcommand{\xmark}{\ding{55}}%

\newcommand{\dataset}{\textsc{Ameli}}
\newcommand{\myTitle}{\dataset{}: Enhancing Multimodal Entity Linking with \\ Fine-Grained Attributes}

 \setlist[itemize]{leftmargin=*}
\setlist[enumerate]{leftmargin=*}

\newpage
\usepackage{appendix}  
\usepackage{diagbox}                

\DeclareMathOperator{\Ab}{\mathbf{A}}

\DeclareMathOperator{\Vb}{\mathbf{V}}

\DeclareMathOperator{\Abbar}{\bar{\Ab}}

\DeclareMathOperator{\Vbbar}{\bar{\Vb}}

\title{\myTitle}

\author{Barry Menglong Yao$^{\spadesuit}$ \quad Sijia Wang$^{\spadesuit}$ \quad Yu Chen$^{\heartsuit}$ \quad Qifan Wang$^{\heartsuit}$  \quad Minqian Liu$^{\spadesuit}$ \\ \textbf{Zhiyang Xu}$^{\spadesuit}$ \quad \textbf{Licheng Yu}$^{\heartsuit}$ \quad \textbf{Lifu Huang}$^{\spadesuit}$ \\
  $^{\spadesuit}$Virginia Tech \quad $^{\heartsuit}$Meta AI \\
  \texttt{\{barryyao,sijiawang,minqianliu,zhiyangx,lifuh\}@vt.edu} \\
  \texttt{\{hugochen,wqfcr,lichengyu\}@meta.com} \\
  }

\begin{document}
\maketitle

\begin{abstract}
We propose attribute-aware multimodal entity linking, where the input consists of a mention described with a text paragraph and images, and the goal is to predict the corresponding target entity from a multimodal knowledge base (KB) where each entity is also accompanied by a text description, visual images, and a collection of attributes that present the meta-information of the entity in a structured format. 
To facilitate this research endeavor, we construct \dataset, encompassing a new multimodal entity linking benchmark dataset that contains 16,735 mentions described in text and associated with 30,472 images, and a multimodal knowledge base that covers 34,690 entities along with 177,873 entity images and 798,216 attributes. To establish baseline performance on \dataset, we experiment with several state-of-the-art architectures for multimodal entity linking and further propose a new approach that incorporates attributes of entities into disambiguation. Experimental results and extensive qualitative analysis demonstrate that extracting and understanding the attributes of mentions from their text descriptions and visual images play a vital role in multimodal entity linking. To the best of our knowledge, we are the first to integrate attributes in the multimodal entity linking task\footnote{The programs, model checkpoints, and the dataset are publicly available at \url{https://github.com/VT-NLP/Ameli}.}. 
 
\end{abstract}

\section{Introduction}

Entity linking aims to disambiguate and link entity mentions within a text to their corresponding entities in knowledge bases. While earlier research~\cite{Onoe2020,Zhang2022,Sun0008,Tang2021,Yang271,ganea-hofmann-2017-deep,Prabhakar3432,Ayoola3893,Ayoola0809854} predominantly focus on linking entities based on text, recent studies have started to extend it to multi-modality where both mentions and entities in knowledge bases are described with text and visual images~\cite{Attention-Based,Weibomel,MultimodalEntity,fasterzs, VTKEL,wikidiverse,TwitterMEL, wang2023benchmarking}. 
However, all these studies view each entity in the knowledge base as an atomic symbol 
while ignoring the meta-information, such as various attributes of each entity, which, we argue, is especially important in disambiguating entities in a multimodal context.

\begin{figure*}[t]
    \centering
    \includegraphics[width=\linewidth]{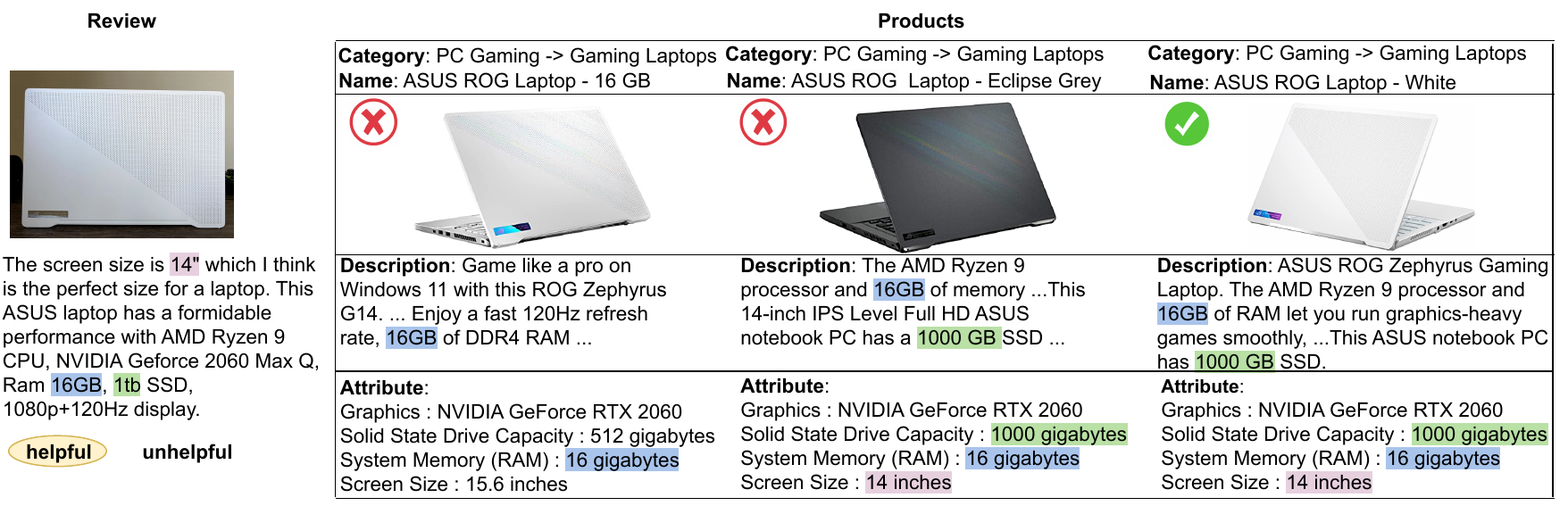}
    \caption{An example for our attribute-aware multimodal entity linking. Left: review text and image; Right: product title, image, description, and attributes. To link the mention \textit{ASUS laptop} to the target entity, we need to be aware of the attributes, e.g., memory and SSD capacity, and image features, e.g., color. 
    }
    \label{fig:example}
    \vspace{-4mm}
\end{figure*}

In this work, we focus on multimodal entity linking (MEL) which requires understanding fine-grained attributes of mentions from both text and images and linking them to the corresponding entities in a target multimodal knowledge base where each entity is also illustrated with text, images, and a set of attributes. Figure~\ref{fig:example} shows an example where each entity, such as \textit{ASUS ROG Laptop - White} in the target knowledge base is described with a set of attributes, such as \textit{Screen Size}, \textit{System Memory}, \textit{Graphics}, and to disambiguate and link a particular mention, e.g., \textit{ASUS laptop} to the target entity, we need to carefully detect the attributes of the mention from its text and image descriptions and compare it against each entity. Such attribute-aware multimodal entity linking is critical to E-commerce domains, e.g., analyzing user opinions from social media posts about particular products. Yet, it is relatively less studied in the 
entity linking literature. 

To support research toward attribute-aware multimodal entity linking, we introduce \dataset{}, which consists of (1) a multimodal knowledge base that includes 34,690 product entities collected from the Best Buy\footnote{\url{https://www.bestbuy.com/}} website and each entity is described with a product name, a product description, a set of attribute categories and values, e.g., ``Color: Black'', and several images; and (2) a multimodal entity linking benchmark dataset that contains 16,735 data instances while each instance contains a text description for a particular entity mention and several images. The goal is to interpret the multimodal context and attributes of each mention and map it to a particular entity in the multimodal knowledge base. 
\dataset{} is challenging as many entities in the knowledge base are about similar products with subtle differences in a few attributes, and thus, the model needs to correctly detect all the attributes from the multimodal context of each mention in order to link it to the target entity.

We conduct baseline experiments with several entity linking methods and propose a new framework consisting of a Natural Language Inference (NLI) based text disambiguation model to compare the mention description and attributes of candidate entities from the knowledge base and an image disambiguation model based on contrastive learning. Though our proposed approach significantly outperforms all the strong baselines, the experimental results still show a large gap between machine (51.51\% F-score) and human performance (74.0\% F-score). The contributions of this work can be summarized as follows:


\begin{itemize}
\item To the best of our knowledge, \dataset{} is the first benchmark dataset 
to support attribute-aware multimodal entity linking, and we are the first to integrate attribute features to improve the multimodal entity linking task. 
\item We propose a new disambiguation approach that considers the multimodal context of mentions as well as attributes of candidate entities, which significantly outperforms all the previous strong baselines on \dataset{}. Ablation studies further demonstrate the benefit and necessity of incorporating attribute information for multimodal entity linking.
\end{itemize}

\section{Related Work}
 \begin{table*}[!htp]
 
\small
\centering
\resizebox{2\columnwidth}{!}{%
\begin{tabular}{l|c|c|c|c|c }
\toprule
\diagbox{\textbf{Dataset}}{\textbf{Feature}}&\textbf{Attribute} & \textbf{Mention Images} &  \textbf{Mention Text} & \textbf{Entity Images} &\textbf{Entity Text} \\
 \midrule
  ~\newcite{Weibomel}  & \xmark & \cmark  & \cmark&  \cmark&  \cmark \\
Wikidiverse~\cite{wikidiverse}  & \xmark & \cmark  & \cmark&  \cmark&  \cmark \\
WIKIPerson~\cite{10.48550/arxiv.2211.04872} & \xmark & \cmark  & \xmark&  \cmark&  \cmark \\
 OVEN-Wiki~\cite{10.48550/arxiv.2302.11154}  & \xmark & \cmark  & \xmark&  \cmark&  \cmark
 \\
ZEMELD ~\cite{fasterzs}  & \xmark & \cmark  & \cmark&  \cmark&  \cmark \\
 MEL\_Tweets~\cite{TwitterMEL}  & \xmark & \cmark  & \cmark&  \cmark&  \cmark \\
M3EL \cite{mel}  & \xmark & \cmark  & \cmark&  \cmark&  \cmark \\
Weibo \cite{Attention-Based}  & \xmark & \cmark  & \cmark&  \cmark&  \cmark \\
 SnapCaptionsKB~\cite{ZeroshotNoisySOcialMedia}  & \xmark & \cmark  & \cmark&  \cmark&  \cmark \\
 VTKEL ~\cite{VTKEL}  & \xmark & \cmark  & \cmark&  \xmark&  \cmark \\
     \newcite{guo2018robust}  & \xmark & \xmark  & \cmark&  \xmark&  \cmark \\
    Zeshel ~\cite{logeswaran2019zero}  & \xmark & \xmark  & \cmark&  \xmark&  \cmark \\
\midrule
\textbf{\dataset{} (Ours)} & \cmark & \cmark & \cmark&  \cmark  &  \cmark 
\\ 

\bottomrule
\end{tabular}%
}
 
\caption{Comparison between \dataset{} and other related datasets. }
\vspace{-4mm}
\label{tab:compare}
 
\end{table*}

 
Previous research on textual entity linking has established various benchmark datasets ~\cite{guo2018robust,logeswaran2019zero,hoffart2011robust,cucerzan-2007-large,milne2008learning} and state-of-the-art neural models~\cite{wu2019scalable,logeswaran2019zero,10.48550/arxiv.2207.04108,peters2019knowledge,ganea2017deep,kolitsas2018end,10.48550/arxiv.2109.03792,10.48550/arxiv.2202.13404,10.48550/arxiv.2010.00904,de2022multilingual}. However, these approaches cannot be directly adapted to multimodal entity linking due to the fundamental differences in input modalities and challenges. 

Multimodal entity linking has recently been explored in various contexts such as social networks \cite{Attention-Based,ZeroshotNoisySOcialMedia,Weibomel,MultimodalEntity}, domain-specific videos \cite{animal} and general news domains \cite{wikidiverse}. These studies focus on reducing noise in the abundant visual input of social networks \cite{Attention-Based,MultimodalEntity}, learning distinguishable entity representations by contrastive learning~\cite{wikidiverse,ZeroshotNoisySOcialMedia,mel}, or directly generating target entity names~\cite{wang2023benchmarking,shi2023generative}.  
Compared to these studies, our research considers the unique attributes along with visual and textual inputs. Table~\ref{tab:compare} compares \dataset{} with other existing entity linking datasets.



Many studies have been proposed to extract attribute values of products from their titles and descriptions by formalizing it as a sequence tagging task~\cite{10.48550/arxiv.2106.02318,10.1145/3219819.3219839,10.18653/v1/p19-1514} or a question-answer problem~\cite{10.1145/3488560.3498377, 10.1145/3394486.3403047}.
Several recent studies~\cite{10.1145/3447548.3467164,10.48550/arxiv.2009.07162,wang2022smartave} incorporate visual clues, such as product images or visual objects, into textual descriptions and extract attribute values based on their fused representations. 
In this study, we explore the potential of leveraging attribute values extracted from noisy user reviews to improve multimodal entity linking and achieve this by implicitly inferring attribute values through Natural Language Inference (NLI).

\section{Dataset Construction}
\paragraph{Data Source}
Our goal is to build (1) a multimodal knowledge base where each entity is described with text, images, and attributes, and (2) an entity linking benchmark dataset where each mention in a given context is also associated with text and several images and can be linked to a specific entity in the multimodal knowledge base. To construct these two benchmark resources, we use Best Buy\footnote{\url{https://www.bestbuy.com/}}, a popular retailer website for electronics such as computers, cell phones, appliances, toys, etc., given that it consists of both multimodal product descriptions organized in a standard format and user reviews in both text and/or images which can be further used to build the entity linking dataset. As shown in Figure~\ref{fig:example}, each product in Best Buy is described with a \textit{product name}, a list of \textit{product categories}, a \textit{product description}, a set of \textit{ attribute categories and values} as well as several \textit{images}\footnote{For simplicity, we show one image for each review or product in the figure, but there could be multiple associated images for both of them.}. Additionally, users can post reviews in text and/or images under each product, while each review can be rated as helpful or unhelpful by other users. We develop scripts based on \texttt{Requests}\footnote{\url{https://requests.readthedocs.io/en/latest/}} to collect all the above information. Each product webpage also requires a button click to display the attributes, so we further utilize \texttt{Selenium}\footnote{\url{https://www.selenium.dev/}} to mimic the button click and collect all the attributes and values for each product. In this way, we collect 38,329 product entities and 6,500,078 corresponding reviews.

\paragraph{Data Preprocessing}\label{sec:preprocess}

\begin{figure}[t]
    \centering
    \includegraphics[width=0.5\textwidth]{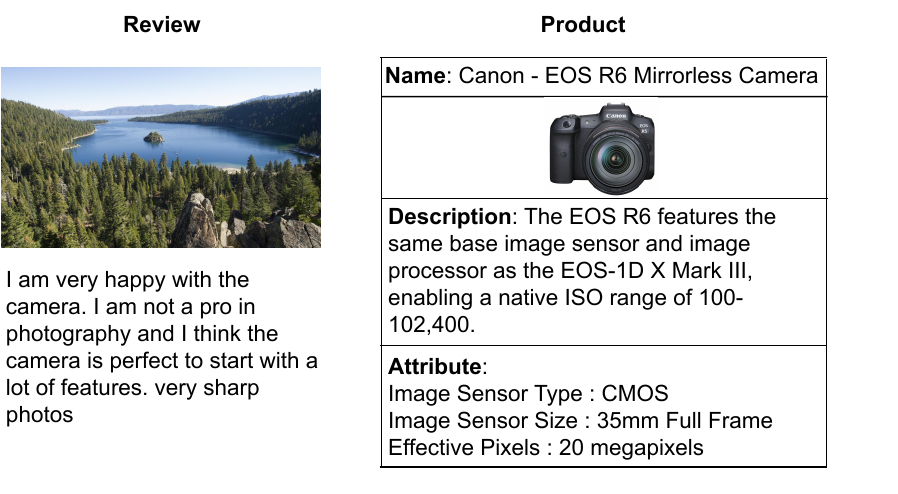}
    \vspace{-8mm}
    \caption{Example of Uninformative Reviews. 
    }
    \label{fig:low_info_review}
    \vspace{-4mm}
\end{figure}

Many reviews are not suitable for the multimodal entity linking task for various reasons. Considering this, we designed several rules to preprocess the collected reviews:
(1) Remove reviews and products without images; (2) Remove reviews with more than 500 tokens, since most of the state-of-the-art pretrained language models can only deal with 512 tokens; (3) Remove a review if it is only labeled as ``unhelpful'' by other users since we observe that these reviews usually do not provide much meaningful information; (4) Validate the links between reviews and their corresponding products and remove the invalid links. There are invalid links because Best Buy links each review to all variants of the target product. For example, for the review of \textit{ ASUS laptop} shown in Figure~\ref{fig:example}, the target product \textit{ ASUS ROG laptop - White} has several other variants in terms of color, memory size, processor model, etc., while Best Buy links the review to all variants of the target product. Since we take each product variant as an entity in our multimodal knowledge base, we detect valid links between reviews and product variants based on a field named \textit{productDetails}, which reveals the gold target product variant information of the review in Best Buy's search response. After obtaining the valid link for each review, we remove invalid links between this review and all other products. 
(5) Remove truncated images uploaded by users as these images cause ``truncated image error'' during loading with standard image tools such as Pillow\footnote{\url{https://pillow.readthedocs.io/en/stable/installation.html}}. (6) Remove reviews containing profanity words based on the block word list provided by \textit{Free Web Header}\footnote{\url{https://www.freewebheaders.com/bad-words-}\\\url{list-and-page-moderation-words-list-for-facebook}}. (7) Review images can also contain irrelevant objects or information; for example, a review image for a fridge can also contain much information on the kitchen. We apply the object detection model~\cite{liu2023grounding} to detect the corresponding object using the entity name as prompt and save the detected image patch as the cleaned review image. We remove an image if the entity object can not be detected from it. Both original images and cleaned images are included in our dataset. (8) We also notice that many reviews do not contain enough context information from the text and images to link the product mention to the target product entity correctly. For example, in Figure~\ref{fig:low_info_review}, the target product is a \textit{Canon camera}. However, the review image does not show the camera itself, and the review text does not contain any specific information about the camera. To ensure the quality of the entity linking dataset, we further design a validation approach (explained in Appendix \ref{sec:filter_low_info}) to filter out reviews that do not contain enough context information.    


\paragraph{Mention Detection}
We identify entity mentions from the reviews based on their corresponding products to construct the entity linking benchmark dataset. To achieve this, we design a pipelined approach 
to detect the most plausible product mention from each review. Given a review and its corresponding product, we first extract all product name candidates from the product title and category by obtaining their root word and identifying a fraction of the root word to be product name candidates with spacy\footnote{\url{https://spacy.io/usage/linguistic-features\#noun-chunks}}. 
For each $n$-gram span ($n\in\{1, 2, 3, 4, 5,6\}$) in the review text, if it or its root form based on lemmatization matches with any of the product name candidates, we will take it as a candidate mention. Each review text may contain multiple mentions of the target product. Therefore, we compute the similarity between each candidate mention and the title of the target product based on SBERT~\cite{reimers-2019-sentence-bert} and choose the one with the highest similarity to be the product mention. This approach achieves a precision of 91.9\% on the detected mentions, based on manual assessment of 200 reviews. Thus, we further apply it to detect product mentions for all reviews and remove the reviews that do not contain product mentions. We then ask one annotator to manually verify and correct all detected mentions in the {\tt Test} set.


\paragraph{Train / Dev / Test Splits}
After all the pre-processing and filtering steps, we obtain 34,690 entities for the multimodal knowledge base and 17,431 reviews for the entity linking benchmark dataset. We name it \dataset{} and split the reviews into training ({\tt Train}), development ({\tt Dev}), and test ({\tt Test}) sets based on the percentages of around 70\%, 10\%, and 20\%, respectively.

\begin{table}[t]

\small
\centering
\begin{tabular}{l|c|c|c  }
\toprule
\textbf{Data} & \textbf{\tt Train} & \textbf{\tt Dev} & \textbf{\tt Test}  \\
 \midrule
\# Reviews  &12,148 & 1,846  &2,741 \\ 
\# Review Images  & 21,780 & 3,369   & 5,323 \\
Avg. \# of Image / Review &1.79 &1.83  &1.94\\
Avg. \# of  Attributes / Review &1.22 &1.62 & 3.54\\
\midrule
\# Products in KB & \multicolumn{3}{c}{  34,690} 
\\
\# Product Images & \multicolumn{3}{c}{ 177,873 } 
\\
\# Product Categories &\multicolumn{3}{c}{986}\\
Avg. \# of Image / Product & \multicolumn{3}{c}{5.13} \\
Avg. \# of Attributes / Product &\multicolumn{3}{c}{23.01}\\

\bottomrule
\end{tabular}%
\caption{Dataset statistics of \dataset{}. 
}
\label{tab:statistic}
\vspace{-5mm}
\end{table}

Note that since we utilize automatic strategies to detect mentions from reviews and filter out uninformative reviews, there is still noise remaining in the \dataset{} though the percentage is low. Thus, we ask humans to verify the {\tt Test} set of \dataset{}. However, it is not trivial for humans to compare each mention with thousands of entities in the target knowledge base. To facilitate entity disambiguation by humans, for each review, we design two strategies to automatically retrieve strong negative candidate entities from the knowledge base: (1) as we know the target product of each review, we first retrieve the top-$N$\footnote{We set $N=10$ as we observe that the top-$10$ retrieved candidates have covered the most confusing negative entities.} most similar entities to the target project from the KB as negative candidates. Here, the similarity between two products is computed based on the cosine similarity scores of their title representations produced by SBERT~\cite{reimers-2019-sentence-bert}; (2) Similarly, we also retrieve the top-$N$ similar entities to the target product based on the cosine similarity scores of their image representations learned by CLIP~\cite{radford2021learning}.
We combine these $2N$ negative candidates together with the target product entity as the set of candidate entities for each review and ask 12 annotators to choose the most likely target entity. Most annotators reach an accuracy of around 80\%, while the overall accuracy is 79.73\%, as shown in Table ~\ref{tab:annotation} in Appendix ~\ref{sec:annotator}. We remove the review if any of the annotators cannot select the target entity correctly. In this way, we obtain 2,741 reviews as the {\tt Test} set. For each review in the {\tt Test} set, we further ask one annotator to label all the attributes ({\tt Gold Attributes}) of each mention.
Table~\ref{tab:statistic} shows the detailed statistics for each split of \dataset{}. Table~\ref{tab:category} in Appendix~\ref{sec:category} 
shows the category distribution of products in the multimodal knowledge base of \dataset{}.

\section{Approach}  
\subsection{Problem Formulation}
\label{sec:approach}

We formulate the task as follows: given a user review $r$ consisting of a text $t_r$, several images $\Vbbar_r=\{v^0_r,...,v^q_r \}$, and an entity mention $m_r$, e.g., ``coffee maker'',  
we aim to link the mention to an unique entity in the target knowledge base (KB).  Each entity $e_j$ in the KB is described with a text description $d_{e_j}$, a title $\hat{t}_{e_j}$, several images $\Vbbar_{e_j}=\{v^0_{e_j},...,v^h_{e_j}\}$
, and a set of attributes $\Abbar_{e_j}=\{a^0_{e_j},...,a^s_{e_j}\}$. Note that the entity title is also one of the attributes. Following previous work~\cite{10.3233/sw-222986}, we solve this task through a two-step pipeline: \textit{Candidate Retrieval}, which retrieves top-$K$ candidate entities $\{e_0,...,e_K \}$  from the KB, and \textit{Entity Disambiguation}, which selects the 
gold entity $e^+$ from the $K$ candidates $\{e_0,...,e_K \}$. Note that $e^+$ may not be in $\{e_0,...,e_K \}$ due to the retrieval error.

\subsection{Candidate Retrieval}
\begin{figure*}[t]
    \centering
    \includegraphics[width=\textwidth]{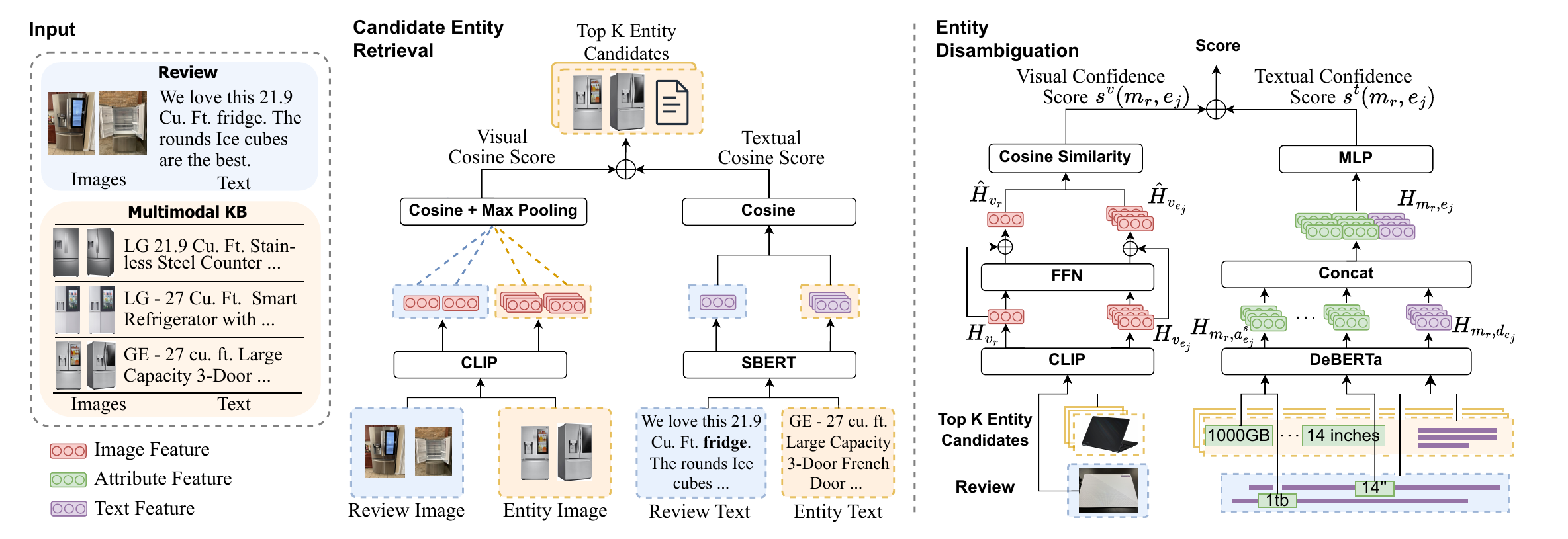}
    \caption{
    Candidate retrieval and entity disambiguation pipeline. 
    We first retrieve the most relevant candidates using cosine similarity with regard to both textual descriptions and images and then predict the gold entity with the NLI-based text disambiguation and contrastive-learning-based image disambiguation modules.
    }
    \label{fig:ret}
    \vspace{-4mm}
\end{figure*}

As shown in Figure \ref{fig:ret}, we retrieve a set of candidate entities from the  KB 
based on textual and visual similarity. For efficiency purposes, we aim to first generate a lookup embedding for each review and entity based on their textual descriptions and visual images, so that the representations can be cached to enable efficient retrieval. 
\paragraph{Text Cosine Similarity}
We apply SBERT \cite{devlin2018bert} to take the review text $t_r$ and entity text $t_{e_j}$\footnote{We append entity title, description, and attributes as the entity textual information for candidate retrieval phase because this combination achieves better performance than other combinations in our preliminary experiments, as shown in Table~\ref{tab:preliminary} in Appendix~\ref{sec:preliminary}.} as input, respectively, and output their representations $\boldsymbol{T}_r$ and $\boldsymbol{T}_{e_j}$\footnote{We use bold symbols to denote vector representations.} based on the {\tt CLS} token. Then we compute a textual cosine similarity score $s^R_t(m_r,e_j)$ for each pair.
\begin{equation}
s^R_t(m_r,e_j)= \text{cosine} (\boldsymbol{T}_{r},\boldsymbol{T}_{e_j})
\end{equation}
The SBERT model is fine-tuned based on the InfoNCE loss~\cite{van2018representation}:
\begin{equation}
\resizebox{0.85\columnwidth}{!}{
$\mathcal{L}(\boldsymbol{T}_r, \boldsymbol{T}_{e^+}, \mathcal{T}) =
-\log\frac{\exp[\text{cos}(\boldsymbol{T}_r, \boldsymbol{T}_{e^+})]}{\sum_{\boldsymbol{T}_{e_j}\in\mathcal{T}} \exp[\text{cos}(\boldsymbol{T}_r,\boldsymbol{T}_{e_j})]}$
}
\end{equation}
where $e^+$ is the {gold} entity of mention $m_r$, 
$\mathcal{T}$ is text representations of candidate entities for $m_r$, including the gold entity $e^+$, standard negative entities whose product categories are different from the gold entity, and in-batch negative entities that are candidate entities 
to other reviews in the same batch\footnote{We remove any duplicate sentences within the same batch}.

\paragraph{Image Cosine Similarity}
To incorporate visual similarity, we employ CLIP \cite{radford2021learning} to obtain image representations, followed by a cosine similarity computation. Since multiple images exist for each review and entity, the CLIP model is fine-tuned based on the InfoNCE loss computed for each review image.
\begin{equation}
\resizebox{0.85\columnwidth}{!}{
$\mathcal{L}(\boldsymbol{V}^q_r, \boldsymbol{V}^h_{e^+}, \mathcal{T}) =
-\log\frac{\exp[\text{cos}(\boldsymbol{V}^q_r, \boldsymbol{V}^h_{e^+})]}{\sum_{\boldsymbol{V}^i_{e_j}\in\mathcal{T}} \exp[\text{cos}(\boldsymbol{V}^q_r,\boldsymbol{V}^i_{e_j})]}$
}
\end{equation}
where $v^q_r$ is one review image, $v^i_{e_j}$ is one entity image, and  $\mathcal{T}$ is image representations of candidate entities for $m_r$, including the gold entity $e^+$, standard negative entities whose product categories are different from the gold entity, and in-batch negative entities.  
The image cosine similarity score between mention $m_r$ and entity $e_j$ is the maximum cosine similarity between their image sets  $\Vbbar_r=\{v^0_r,...,v^q_r \}$ and $\Vbbar_{e_j}=\{v^0_{e_j},...,v^h_{e_j}\}$.
\begin{gather}
\small
s^R_v(m_r,e_j)=   \smash\displaystyle\max_{v^q_r\in \Vbbar_r,v^i_{e_j}\in \Vbbar_{e_j}}\text{cosine} (\boldsymbol{V}^q_r,\boldsymbol{V}^i_{e_j})
\end{gather}

\paragraph{Candidate Selection} A weighted sum is applied to the textual and visual cosine similarity scores to obtain the merged similarity scores $s^R(m_r,e_j)$, based on which, we select the top-$K$ ranked entities as candidates.
\begin{gather}
\small
\resizebox{0.85\columnwidth}{!}{
$s^R(m_r,e_j)=   \lambda  \cdot s^R_t(m_r,e_j)+(1-\lambda) \cdot s^R_v(m_r,e_j)$
}
\end{gather}
where $\lambda$ is a coefficient searched on the {\tt Dev} set.


\subsection{Entity Disambiguation} \label{sec:entity_disambiguation}
 As shown in the right part of Figure \ref{fig:ret}, our disambiguation model comprises an NLI-based text module and a contrastive learning-based image module.
 
\paragraph{Preprocess} We first apply four methods to extract attributes for each mention from its review text and images: 
(1) \textit{OCR}: since there may exist text inside review images such as brand names, we apply an off-the-shelf OCR tool\footnote{\url{https://github.com/JaidedAI/EasyOCR}} to recognize texts within each review image;
(2) \textit{String Match}, that identifies attribute values from review text based on the attribute values of the top-$K$ retrieved candidate entities, e.g., if a candidate entity has an attribute value like ``16 gigabyte'' which also occurs in the review text, we will take it as a value for the attribute category ``Memory''; 
(3) \textit{ChatGPT}~\cite{chatgpt}: in many cases, the review may contain the attribute value which is described in a slightly different form, such as ``16 GB'', which cannot be identified by \textit{String Match}. 
So, we further leverage ChatGPT and formalize our attribute value extraction task as a multiple-choice QA~\cite{robinson2022leveraging} problem by treating each attribute category as a question and the corresponding attribute values from the top-$K$ candidate entities as options, as detailed in Figure \ref{fig:mcqa_prompt} in Appendix \ref{sec:prompt}. Limited by the computational cost, we apply ChatGPT on 11 common attribute categories, including ``\textit{Brand}, \textit{Color}, \textit{Model Number}, \textit{Product Title}, \textit{Screen Size}, \textit{Processor Brand}, \textit{Processor Model}, \textit{System Memory (RAM)}, \textit{Graphics}, \textit{Solid State Drive Capacity}, \textit{GPU Brand}''.
(4) \textit{GPT-2}~\cite{radford2019language}: for attribute categories not covered by \textit{ChatGPT} method, we further apply GPT-2 to generate attribute values for all attribute categories in a zero-shot text completion manner, with the prompt template ``{\tt Attribute Value Extraction:\textbackslash n \#Review\_text \textbackslash n \#Attribute\_key:}'', where {\tt Attribute Value Extraction} is the text prompt, {\tt \#Review\_text} is the corresponding review text and {\tt \#Attribute\_key} is the attribute to be extracted, as shown in Figure \ref{fig:completion_prompt} in Appendix \ref{sec:prompt}. For all the approaches discussed above, we only keep the attribute values that match any attribute value of top-$K$ candidate entities. The resulting attribute value set is denoted as {\tt System Attribute}. We then filter out candidate entities whose attribute values do not match the {\tt System Attribute} of each mention. Since we don't manually label the attributes of mentions in the {\tt Train} and {\tt Dev} datasets, we clean the {\tt System Attribute} to obtain {\tt Gold Attribute} by removing the attributes that do not match with the attributes of the gold entity product.


\paragraph{Text-based Disambiguation}
Our text-based disambiguation module is based on Natural Language Inference (NLI) with the motivation that the review text should imply the product attribute if it is mentioned in the review. For example, given the review ``\textit{I was hoping it would look more pink than it does, it's more of a gray-toned light pink. Not a dealbreaker. I still like this bag}'', it should imply the attribute value of the target product, e.g., ``\textit{The color of this bag is pink}'', while contradicting attribute values of other products, e.g., ``\textit{The color of this bag is black}''. 
Thus, for each review with a mention $m_{r}$ and text $t_r$, given a set of candidate entities $\{e_0,...,e_j\}$ with their descriptions $\{d_{e_0},...,d_{e_j}\}$ and attribute values $\{a^0_{e_0},...,a^0_{e_j}\}$, ..., $\{a^s_{e_0},...,a^s_{e_j}\}$, as there could be many attributes of candidate entities that are not mentioned in the review, we first select a subset of attribute values for the candidate entities based on the attribute categories covered in the {\tt System Attribute} of mention $m_{r}$. 
Then, we pair each entity attribute or the entity description with the review description and feed each pair into a DeBERTa~\cite{he2023debertav3} encoder, which is fine-tuned and has shown promising performance on general NLI tasks, to obtain their contextual representations 
\begin{gather}
\small
H_{m_{r},d_{e_j}}=\text{DeBERTa}(d_{e_j},[m_r:t_r])\\
\small
H_{m_{r},a^s_{e_j}}=\text{DeBERTa}([m_r:t_r],a^s_{e_j}) 
\end{gather}
where $:$ denotes the concatenation operation. For each entity with multiple attribute values, we concatenate all the contextual representations obtained from DeBERTa and feed it through MLP to predict the final NLI score: 
\begin{gather}
\small
\resizebox{0.85\columnwidth}{!}{
$H_{m_r,e_j}=[H_{m_{r},a^0_{e_j}}: H_{m_{r},a^1_{e_j}}...,H_{m_{r},a^s_{e_j}}:H_{m_{r},d_{e_j}}]$
}
\\
\small
\resizebox{0.51\columnwidth}{!}{
$s^t(m_r,e_j)=\text{MLP}(H_{m_r,e_j})$
}
\end{gather}

During training, we optimize the text-based disambiguation module based on the cross-entropy objective:
\begin{equation}
\small
\mathcal{L}^t(m_r,e^+) =-\log\frac{\exp(s^t(m_r,e^+))}{\sum_{j=0}^{K-1} \exp(s^t(m_r,e_j))}
\end{equation}
where $e^+$ is the gold entity, and $K$ is the number of retrieved candidate entities. 

\paragraph{Image-based Disambiguation}
Given the review image $v_r$\footnote{Following~\cite{wikidiverse,10.48550/arxiv.2211.04872}, we select one image for each review and entity during the disambiguation, based on the cosine similarity score of their CLIP representations,
which also showed better performance than using all images in our preliminary experiments. 
}
 and entity images for a set of candidate entities $\{v_{e_0},...,v_{e_j}\}$, we feed them into CLIP to obtain their image representations $\{H_{v_r}, H_{v_{e_0}},...,H_{v_{e_j}} \}$. Inspired by previous studies~\cite{zhang2022tip,10.48550/arxiv.2110.04544,10.48550/arxiv.2211.04872}, we feed these through 
 a feed-forward layer and residual connection to adapt the generic image representations to a task-oriented semantic space 
\begin{gather}
\small
\hat{H}_{v_{e_j}}=H_{v_{e_j}}+\text{ReLU}(H_{v_{e_j}}\cdot W^e_1)\cdot W^e_2 \\
\small
\hat{H}_{v_r}=H_{v_r}+\text{ReLU}(H_{v_r}\cdot W^r_1)\cdot W^r_2 
\end{gather}
where $W^r_1$and $W^r_2$ are learnable parameters for review representation learning, $W^e_1$and $W^e_2$ are learnable parameters for entity representation learning. 

We apply the following contrastive loss during training based on the cosine similarity scores. 
\begin{gather}
\small
\mathcal{L}^v(m_r, e^{+}) =
-\log\frac{\exp(\text{cos}(\hat{\boldsymbol{H}}_{v_r}, \hat{\boldsymbol{H}}_{v_{e^+}}))}{\sum_{e_j\in B} \exp(\text{cos}(\hat{\boldsymbol{H}}_{v_r},\hat{\boldsymbol{H}}_{v_{e_j}}))}
\end{gather}
where $B$ is the set of all entities in the current batch since we utilize in-batch negatives to improve our model's ability to distinguish between gold and negative entities. 

\paragraph{Inference }
During inference, we combine the NLI score $s^t(m_r,e_j)$ from the text-based disambiguation module, the cosine similarity score $s^v(m_r,e_j)$ from the image-based disambiguation module and the retrieval score from the candidate retrieval model, and predict the entity with the highest weighted score $s(m_r,e_j)$ as the target
\begin{equation}
\small
s^v(m_r,e_j)=\text{cos}(\hat{\boldsymbol{H}}_{v_r}, \hat{\boldsymbol{H}}_{v_{e_j}})
\end{equation}
\begin{equation}
\begin{aligned}
\small 
s(m_r,e_j) &=\lambda_1s^t(m_r,e_j)+\lambda_2s^v(m_r,e_j) \\
& +(1-\lambda_1-\lambda_2)s^R(m_r,e_j)
\end{aligned}
\end{equation} 
where $\lambda_1,\lambda_2$ are coefficients tuned on the {\tt Dev} set. 
\begin{table*}[!htp]
\small
\centering
\footnotesize
\begin{tabular}{l|c|c|  c| c | c|c}
\toprule
\textbf{Modality}  &     \textbf{Method}     &\textbf{  Recall@1} &    \textbf{  Recall@10}&\textbf{  Recall@20}&\textbf{  Recall@50} &\textbf{  Recall@100}  \\
\midrule
T & Pre-trained SBERT     & 19.52&  46.63& 57.06& 71.18&  82.52\\
V & Pre-trained CLIP      & 14.45& 39.00& 47.25&59.25&68.77\\ 
T+V & Pre-trained CLIP/SBERT       &27.14  & 59.76 &67.75&  77.49 &82.60  \\
\midrule
T & Fine-tuned SBERT   &32.65 & 66.65&76.87& 87.34&93.32\\
V & Fine-tuned CLIP  &28.06  &62.82&  71.76& 80.48&86.25\\
T+V& Fine-tuned CLIP/SBERT &\textbf{48.12}  & \textbf{85.84}&\textbf{90.26}&\textbf{93.69}& \textbf{95.11}\\
\bottomrule
\end{tabular}%
\vspace{-2mm}
\caption{Performance of candidate retrieval.  The modality of T and V represents the textual context and visual context, respectively. 
}
 \label{tab:candidate_retrieval}
\vspace{-2mm}
\end{table*}
\begin{table*}[ht]
\small
\centering
\begin{tabular}{l|c| c| c|c}
\toprule
\textbf{Modality}&\textbf{w Attribute}&\textbf{Method}&  \textbf{Disambiguation F1  (\%)} & \textbf{End-to-End F1  (\%)} \\ 
\midrule

-&No&Random Baseline   &10.00 & 8.58  \\ 
V&No&V2VEL~\cite{sun2022visual}   &  19.55& 16.71 \\ 
T&No&V2TEL~\cite{sun2022visual}  &  19.68& 17.07  \\ 
T+V&No&V2VTEL~\cite{sun2022visual}   &  31.36   &   30.35  \\ 
T+V &No&LLaVA~\cite{liu2023visual}  & 23.33 &20.03  \\
T+V &No&GHMFC~\cite{10.1145/3477495.3531867}    & 12.15& 12.11\\ 
T+V &Filter&GHMFC*~\cite{10.1145/3477495.3531867}   & 22.74 & 21.78 \\
T+V &No&Wikidiverse~\cite{10.18653/v1/2022.acl-long.328}  & 13.26 &10.76\\
T+V &Filter&Wikidiverse*~\cite{10.18653/v1/2022.acl-long.328}  &24.69 &20.54  \\

\midrule
T+V&No&Our Approach\_w/o\_Attribute     &52.57  &44.84   \\ 
T&System&Our Approach\_w/o\_Image  &  44.50&38.53 \\ 
V&System&Our Approach\_w/o\_Text  & 42.54&36.63  \\ 
T+V&System&Our Approach   & 60.22 &  51.51 \\ 
\midrule
T+V&Gold&Our Approach & 73.57 & 62.82    \\ 
T+V&No&Human & 80.00 &  74.00  \\ 
\bottomrule
\end{tabular}%
\vspace{-2mm}
\caption{\label{tab:widgets2} Performance of entity disambiguation. Gold stands for the \texttt{Gold Attribute} mentioned in the review, System stands for the \texttt{System Attribute} predicted by our methods, while \texttt{Filter} applies a straightforward elimination of candidate entities whose entity attributes do not align with the predicted review attributes.
} 
\label{tab:entity_disambiguation}
\vspace{-3mm}
\end{table*}

\section{Experiments and Analysis}

\subsection{Candidate Retrieval}\label{sec: entity retrieval}
For each review, we retrieve the top-$K$ candidate entities from the target KB and evaluate the retrieval performance based on Recall@$K$ ($K=1,10,20,50,100$). As shown in Table \ref{tab:candidate_retrieval}:  (1) the multimodal retrieval outperforms the single-modality retrieval, demonstrating that both text and image information complement each other. (2) All models have obtained significant improvements after fine-tuning (e.g., Recall@10 improves by 23.3 points on average), which indicates the effectiveness of fine-tuning on our dataset. (3) After fusing image and text cosine similarity scores, our model achieves 95\% of Recall@100, which shows that most relevant entities can be retrieved from the multimodal knowledge base. 

\subsection{Entity Disambiguation}\label{sec: entity disambiguation}
We further evaluate the entity disambiguation performance based on the micro F1-score under the (1) \textit{End-to-End} setting, where models predict the target entity from the top-$K$ ($K=10$) retrieved entities, and (2) \textit{Disambiguation} setting, where models are evaluated on the subset of 2,353 testing instances whose gold entities exist in the top-$K$ ($K=10$) retrieved candidates. We compare our approach with a \textit{Random Baseline} which chooses the target product randomly and several high-performing baselines for multimodal entity linking as detailed in Appendix~\ref{sec:baseline}.


As shown in Table \ref{tab:entity_disambiguation}, our approach outperforms all baseline methods and reaches 51.51\% of \textit{End-to-End} F1 score. One reason for the low performance is the error propagation from the Candidate Retrieval phase to Disambiguation. Our model can reach 60.22\% of F1 score under the \textit{Disambiguation} setting when the gold entity exists in the retrieved candidate set. 

To evaluate the impact of each modality on entity disambiguation, we design ablated models of our approach by removing text, image, or attributes from the model input. The results show that each modality can benefit the disambiguation, while the attribute information brings a considerable performance improvement. A possible reason for this performance gap is that attributes provide a strong, direct signal for the coreference between the review context of each mention and its gold entity. In addition, during the text-based disambiguation, we use \texttt{System Attribute} to select a subset of attribute values for the candidate entities. However, the \texttt{System Attribute} may contain incorrect attributes or miss some attributes of the mention that are also contained in the review. To evaluate its impact on text-based disambiguation, for the {\tt Test} set, we use \texttt{Gold Attribute} labeled by humans, which yields significantly higher F1 scores, e.g., 73.57\% F1 for the \textit{Disambiguation} setting and 62.82\% for the \textit{End-to-End} setting.




Finally, we also set up a human performance for entity disambiguation by randomly sampling 50 reviews, with 10 candidate entities for each review, for the \textit{Disambiguation} setting and 50 reviews for the \textit{End-to-End} setting, and ask two annotators to execute manual entity-linking. Based on Fleiss $\kappa$~\cite{fleiss1971measuring}, the agreement score between the two annotators is 0.69 for the \textit{Disambiguation} setting and 0.71 for the \textit{End-to-End} setting. 
We consider a human prediction accurate only if both annotators provide the true label. As we can see in Table~\ref{tab:entity_disambiguation}, there is a considerable gap between our model and \textit{Human Performance}. 





\section{Remaining Challenges}~\label{sec:challenge} 
We randomly sample 50 reviews linked to incorrect entities under the \texttt{System Attribute}  setting from the \texttt{Test} and identify the following key challenges for the entity disambiguation task\footnote{We analyze the \texttt{System Attribute} setting instead of the \texttt{Gold Attribute} since the gold review attribute may not always be available in the real world application.}.

\paragraph{Attribute Extraction:}
 User reviews often contain informal language, idiomatic expressions, and diverse writing styles. This linguistic variability makes it challenging to accurately extract specific attribute values as different users might use other terms to describe the same attribute. Furthermore, our knowledge base encompasses over 30,000 attribute values. Determining the attribute referenced within a given review poses a challenging inference task. For 10\% of the errors, our method fails to extract some key attributes. For example, given review \#1 \textit{``Plus their are 10 programmable buttons and rated to 50 million clicks with omron swicthes now you can't beat that.''} in Figure~\ref{fig:error_example} in Appendix \ref{sec:error_example},  we can distinguish the gold entity with 10 buttons from the candidate entity with 17 buttons after extracting the attribute ``Number of Buttons (Total): 10''. Recognizing brand logos or integrating a better OCR model to detect text within images will also increase the quality of \texttt{System Attribute}, as shown in review \#2. More analysis on attribute extraction module is detailed in Appendix \ref{sec:attribute_extraction}. 

\paragraph{Reasoning over Attributes:}
18\% of the errors can be fixed if the model pays more attention to appropriate attributes or conducts reasoning based on the attribute. For example, the review \#3 in Figure~\ref{fig:error_example} in Appendix \ref{sec:error_example} claims \textit{`` i bought this because you can use it on your phone too ''}. As a result, we can skip the candidate entity with the attribute  ``Compatible Platform(s): Windows, Mac, PlayStation 4, PlayStation 5'' since it does not support phones. In some cases, \texttt{System Attribute} contains the key review attributes to distinguish the gold entity from the candidate entity. However, the model is fed with abundant multimodal context and fails to focus on the distinguishable attribute. For example, in review \#4, the model fails to take care of the attribute ``Carafe Capacity''.

\paragraph{Fine-grained Image Matching:}
In 32\% of the errors, the gold entity and candidate entity can be distinguished based on fine-grained image texture. For example, in review \#5 in Figure~\ref{fig:error_example} in Appendix \ref{sec:error_example}, the delicate pattern in the computer case acts as the main hint to link to the gold entity. Since these inconspicuous patterns can be pretty elusive to spot, visual attributes will be helpful to guide attention in some cases. For example, in review \#6, the difference between the gold and candidate entities is whether there is an Ice and Water Dispenser on the fridge surface. With the visual attribute ``Ice and Water Dispenser Location: External,'' the model can focus on the image patch on which the Ice and Water Dispenser is normally located.  

\paragraph{Candidate Retrieval:}
26\%  of the disambiguation errors are due to the gold entity not being in the top 10 retrieved candidates, a.k.a. error propagation from the candidate retrieval phase. We notice the following retrieval error patterns by comparing the gold entity with the top 10 retrieved candidates. (1) Similar to the disambiguation phase, attribute-based match and fine-grained image match can help distinguish the gold entity from candidate entities. (2) One unique error in the retrieval phase happens when one of the review images is irrelevant to the gold entity, thus introducing noise when computing the maximum image similarity score.  

\section{Conclusion}
 
We propose attribute-aware multimodal entity linking, which requires features extracted from images, text descriptions, and structured attributes to disambiguate and link each mention to the corresponding entity in a target knowledge base. To support this research, we construct \dataset{}, consisting of a multimodal knowledge base that contains 34,690 product entities described with text, images, and fine-grained attributes, and a multimodal review dataset that contains 16,735 review instances while each review is also associated with a text description and an image. We experiment with several high-performance entity-linking approaches, including a new approach that incorporates attributes of entities for disambiguation. Experimental results show that the attributes indeed significantly enhance the model performance, but still, there is a large gap between the machine and human performance. 






\section*{Limitations}
\paragraph{Advanced Approach to Incorporate Attributes} In this research endeavor, we propose an innovative approach incorporating attributes into the disambiguation process using a Natural Language Inference (NLI)-based framework. However, we acknowledge that this approach may not fully harness the potential of attributes. Attribute-aware encoding ~\cite{wei20212,saini2022disentangling}, attribute-based zero-shot learning~\cite{lampert2013attribute}, and attribute-aware retrieval~\cite{wei20212,dong2023region} can be promising directions for future work.

\section*{Ethics Statement}
 
We carefully follow the ACM Code of Ethics \footnote{\url{https://www.aclweb.org/portal/content/acl-code-ethics}} and have not found potential societal impacts or risks so far. To the best of our knowledge, this work has no notable harmful effects and uses, environmental impact, fairness considerations, privacy considerations, security considerations, or other potential risks. Our dataset does not contain sensitive personally identifiable information such as name, address, or phone number.

Since our dataset contains user reviews, some claims may be offensive and we remove reviews containing profanity words as specified in Sec~\ref{sec:preprocess}.

\section*{Acknowledgements}
This research is based upon work supported by Meta AI.
We also extend our gratitude to Daniel Hajialigol for his assistance with the initial web crawling, to Meghana Holla for her assistance in refining the initial version of this paper, and to Jingyuan Qi, Ying Shen, Zoe Zheng, Pritika Ramu, Samhita Reddivalam, Sai Gurrapu, Indrajeet Kumar Mishra, Mingchen Li, and Mohammad Beigi for their invaluable contributions to dataset annotation and human experimentation.

\bibliography{anthology,section/custom}
\appendix



\section{Filtering of Uninformative Reviews} \label{sec:filter_low_info}
For each review and its corresponding product,  we extract four features, including \textit{\# of mentioned attributes} (i.e., the number of product attributes mentioned in the review based on string match), \textit{image-based similarity} (i.e., the maximum similarity between review images and gold product images based on CLIP~\cite{radford2021learning} image embeddings), \textit{description-based similarity} (i.e., the similarity between gold product description and review text based on SBERT~\cite{reimers-2019-sentence-bert}), \textit{title-based similarity} (i.e., the similarity between the gold product title and review text using SBERT~\cite{reimers-2019-sentence-bert}). 
We further manually annotate 500 pairs of reviews and products while each pair is assigned with a label: \textit{positive} if the review is informative enough to correctly link the mention to the target product, otherwise, \textit{negative}, and use them to evaluate a threshold-based approach which predicts the reviews as uninformative reviews if the four extracted feature scores, \{\textit{\# of mentioned attributes}, \textit{image-based similarity}, \textit{description-based similarity}, \textit{title-based similarity}\},  do not overpass the four corresponding thresholds, which are hyperparameters searched on these examples. The threshold-based method reaches 85\% of precision and 82\% of recall in predicting informative reviews on these 500 examples\footnote{We compared the threshold-based method with a series of classifiers, like SVM, by training these classifiers on 385 examples and testing on 165 examples. Threshold-based Method reaches the highest accuracy.}. We further apply it to clean the dataset by removing the reviews predicted as uninformative. 

\section{Human Annotation}\label{sec:annotator}
\begin{figure*}[t]
    \centering
    \includegraphics[width=\textwidth]{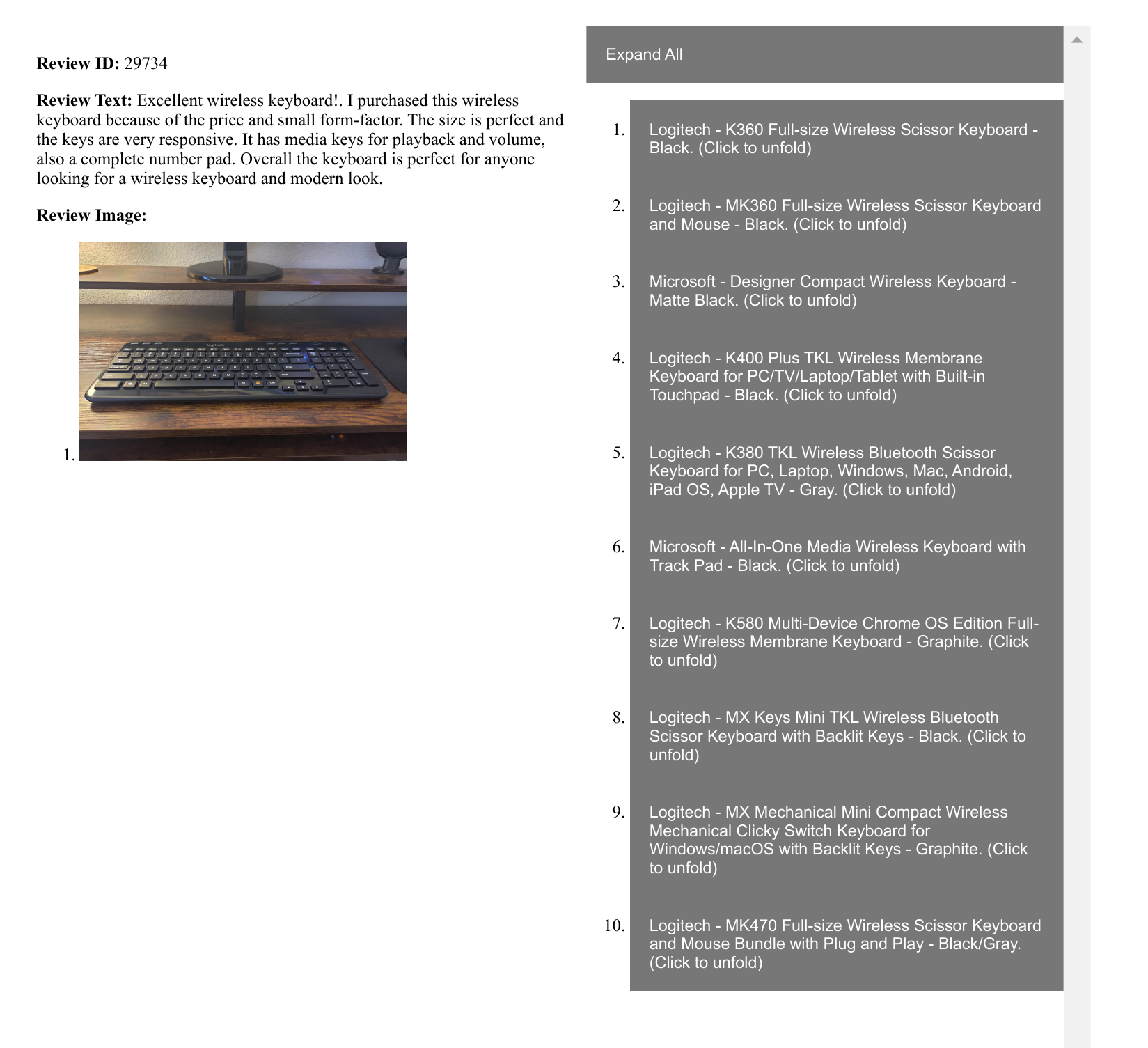}
    \caption{
    Screenshot of human annotation tool.
    }
    \label{fig:annotation}
\end{figure*}

We recruited 12 student volunteers as annotators. 8 of them are from China, and 4 volunteers are from India. For human annotation, we provide the annotation tool as shown in Figure~\ref{fig:annotation} and the following instructions to annotators:
\begin{itemize}  
    \item Open one of your annotation web pages
    \item Click on product 1 to expand its text, images, and attributes information. Compare the review with product 1. Are there any specific product attributes, e.g., memory size, color, can be recognized in the review text/images? Do review images and product images share the same color, shape, or subtle pattern?
    \item If you want to zoom in on any image, you can click on the image, and it will be shown on full screen.
     \item If it seems that product 1 is not the target product, you can fold it by double-clicking the dark product header, and begin to check product 2, product 3, and so on.
     \item Finally, you find the target product. Now you can record the index (1-21) on the provided answer sheet.
\end{itemize}

Table ~\ref{tab:annotation} shows the annotation accuracy for each annotator. 
\begin{table}[ht]

\small
\centering
\resizebox{\columnwidth}{!}{%
\begin{tabular}{l|c|c|c }
\toprule
\textbf{ Annotator ID} & \textbf{ \#Correct} & \textbf{ \#Finished} & \textbf{ Accuracy (\%)} \\
 \midrule
1&244&330&73.94\\ 
2&256&330&77.58\\ 
3&274&330&83.03\\ 
4&240&330&72.73\\ 
5&270&324&83.33\\ 
6&253&330&76.67\\ 
7&124&170&72.94\\ 
8&290&330&87.88\\ 
9&222&330&67.27\\ 
10&295&330&89.39\\ 
11&272&330&82.42\\ 
12&285&330&86.36\\ 
\midrule 
Overall &3025 &3794 &79.73\\ 
\bottomrule
\end{tabular}%
}
\caption{The annotation result. \#Finished stands for the reviews annotated by the corresponding annotator, while \#Correct stands for the correct prediction.} 
\label{tab:annotation}
\vspace{-2mm}
\end{table}

\section{Category Distribution} \label{sec:category}
Table ~\ref{tab:category} shows the category distribution. 
\begin{table}[h]
\centering
\resizebox{0.5\textwidth}{!}{%
\begin{tabular}{l|c| c }
\toprule
\textbf{Category}&\textbf{\# Product}&\textbf{Percentage \%}  \\ 
\midrule
 All Refrigerators&847  &2.44   \\
Action Figures (Toys)&730   &2.10   \\  
Dash Installation Kits&682   &1.97   \\  
Wall Mount Range Hoods&680   &1.96   \\  
 Nintendo Switch Games&628   &1.81  \\  
 Gas Ranges&603   &1.74  \\  
Building Sets \& Blocks (Toys)&576   &1.66   \\  
 Nintendo Switch Game Downloads&574   &1.65  \\  

PC Laptops&554   &1.60   \\  
Cooktops&547  &1.58   \\  
\bottomrule
\end{tabular}%
}
 
\caption{Category Distribution of 10 most frequent categories. \textbf{\# Product} indicates the number of products in the corresponding category while \textbf{Precentage} indicates how many percentages of all products are in this category.
}
\label{tab:category}
\end{table}

\section{Preliminary Experiments}\label{sec:preliminary}

\begin{table*}[t]
\small
\centering
\footnotesize
\begin{tabular}{l|c|c|  c| c | c|c}
\toprule
\textbf{Text Field}  &     \textbf{Method}     &\textbf{  Recall@1} &    \textbf{  Recall@10}&\textbf{  Recall@20}&\textbf{  Recall@50} &\textbf{  Recall@100}  \\
\midrule
 
Title & Pre-trained SBERT    & 12.29& 38.12& 48.92& 63.81&  75.88\\
Desc  & Pre-trained SBERT    & 14.85& 40.42& 50.31&  63.81&   74.17\\
Attri & Pre-trained SBERT    & 12.81& 36.88& 47.46&63.77&  77.67\\
Title+Attri  & Pre-trained SBERT    & 16.42& 43.56&54.10&  67.68&79.53\\
Title+Desc & Pre-trained SBERT    & 19.08&  47.21&  57.10& 69.21&  79.82\\
Attri+Desc & Pre-trained SBERT    & 17.62& 45.06&55.45& 69.06& 80.48\\
Title+Desc+Attri & Pre-trained SBERT    & 19.52&  46.63& 57.06& 71.18&  82.52\\

\bottomrule
\end{tabular}%
\caption{Preliminary performance of entity candidate retrieval based on the cosine similarity between the review text and the corresponding entity text field.  ``Title'', ``Desc'', and ``Attri'' stand for entity title, entity description, and entity attributes, respectively. ``Title+Desc'' stands for the concatenation of the entity title followed by the entity description; the other combinations follow the same order.  
}
 \label{tab:preliminary}
\end{table*}

Table ~\ref{tab:preliminary} shows the preliminary experiments on candidate retrieval.

\section{Prompt Templates for GPT-2, ChatGPT, Vicuna, and LLaVA}\label{sec:prompt} 
\begin{figure*}[t]
    \centering
    \includegraphics[width=0.8\textwidth]{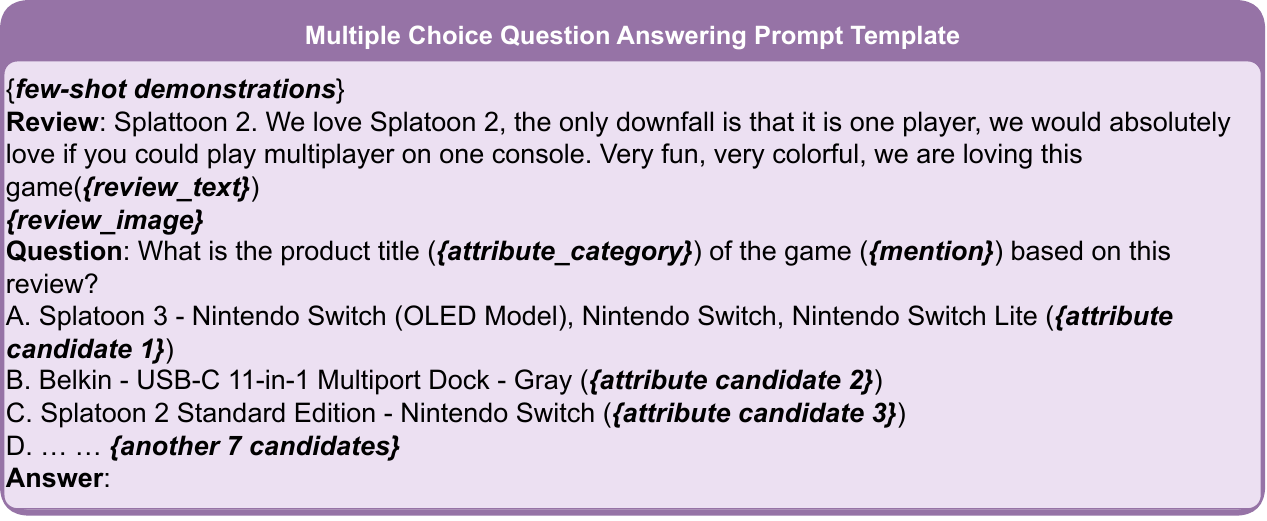}
    \caption{
    The multiple-choice QA prompt template is applied in ChatGPT-based and Vicuna-based Attribute Extraction and LLaVA-based Entity Disambiguation.
    }
    \label{fig:mcqa_prompt}
\end{figure*}

\begin{figure*}[t]
    \centering
    \includegraphics[width=0.9\textwidth]{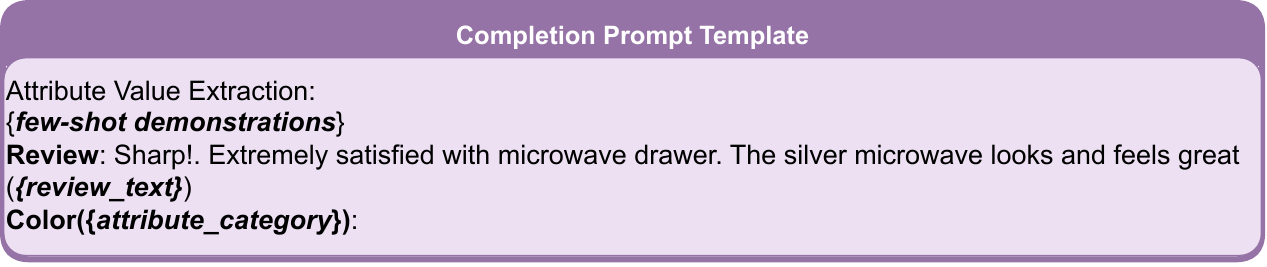}
    \caption{
    The text completion prompt template is applied in GPT-2 based Attribute Extraction.
    }
    \label{fig:completion_prompt}
\end{figure*}

We show the applied prompt templates in Figure \ref{fig:mcqa_prompt} and Figure \ref{fig:completion_prompt}.

\section{Baseline Approaches}\label{sec:baseline}
We compare our approach with several baselines on the entity disambiguation task: 
\begin{itemize}
\item a \textit{Random Baseline} which chooses the target product randomly; 
\item \textit{V2VEL}~\cite{sun2022visual}, which is a visual entity linking model with entity image and mention image as the input, Resnet150~\cite{he2015deep} as the image encoder, and one adapter layer to adapt the representation to the task representation space;
\item \textit{V2TEL}~\cite{sun2022visual},  which incorporates CLIP to encode entity text and mention image for prediction;
\item \textit{V2VTEL}~\cite{sun2022visual}, which combines \textit{V2VEL} and \textit{V2TEL} in a two-step retrieval-then-rerank pipeline. We first apply the trained \textit{V2VEL} model to select top-$L$ entities from top-$K$ candidate entities (K>L), then use the trained \textit{V2TEL} model to predict the gold entity from top-L entities. We set K=10 and L=5 in our experiments.
\item \textit{GHMFC}~\cite{10.1145/3477495.3531867}, which applies textual-guided visual attention and visual-guided textual attention to extract multimodal features, followed by a gated fusion and contrastive training; 
\item \textit{Wikidiverse}~\cite{wikidiverse}\footnote{\textit{V2VEL}, \textit{V2TEL}, \textit{V2VTEL}, \textit{GHMFC}, and \textit{Wikidiverse} are all fine-tuned on our dataset. For a fair comparison, they are used to predict the gold entity from top-$K$ candidate entities, the same setting as our method.}, which concatenates patch-level image representation and token-level text representation and feeds them into a self-attention transformer for multimodal fusion;  
\item \textit{GHMFC*} and \textit{Wikidiverse*}, where we improve \textit{GHMFC}~\cite{10.1145/3477495.3531867} and \textit{Wikidiverse}~\cite{wikidiverse} with a post-process  ``Attribute Filter'', which leverages a straightforward elimination of candidate entities whose entity attributes do not align with the predicted review attributes.
\item \textit{LLaVA}~\cite{liu2023visual}, where we conduct an experiment of employing a SOTA multimodal large model, i.e., LLaVA, directly for attribute-aware multimodal entity linking in a few-shot manner. Specifically, given a particular review and an entity mention, we first employ the same candidate retrieval approach to obtain the  top-$K$ ($K$=10) candidate entities, then we ask the LLaVA model to directly choose the most plausible candidate entity title from the 10 candidate entity titles based on the multiple-choice QA prompt template in Figure \ref{fig:mcqa_prompt}.

\end{itemize}

\section{Error Examples}\label{sec:error_example}
\begin{figure*}[!ht]
    \centering
    \includegraphics[width=\textwidth]{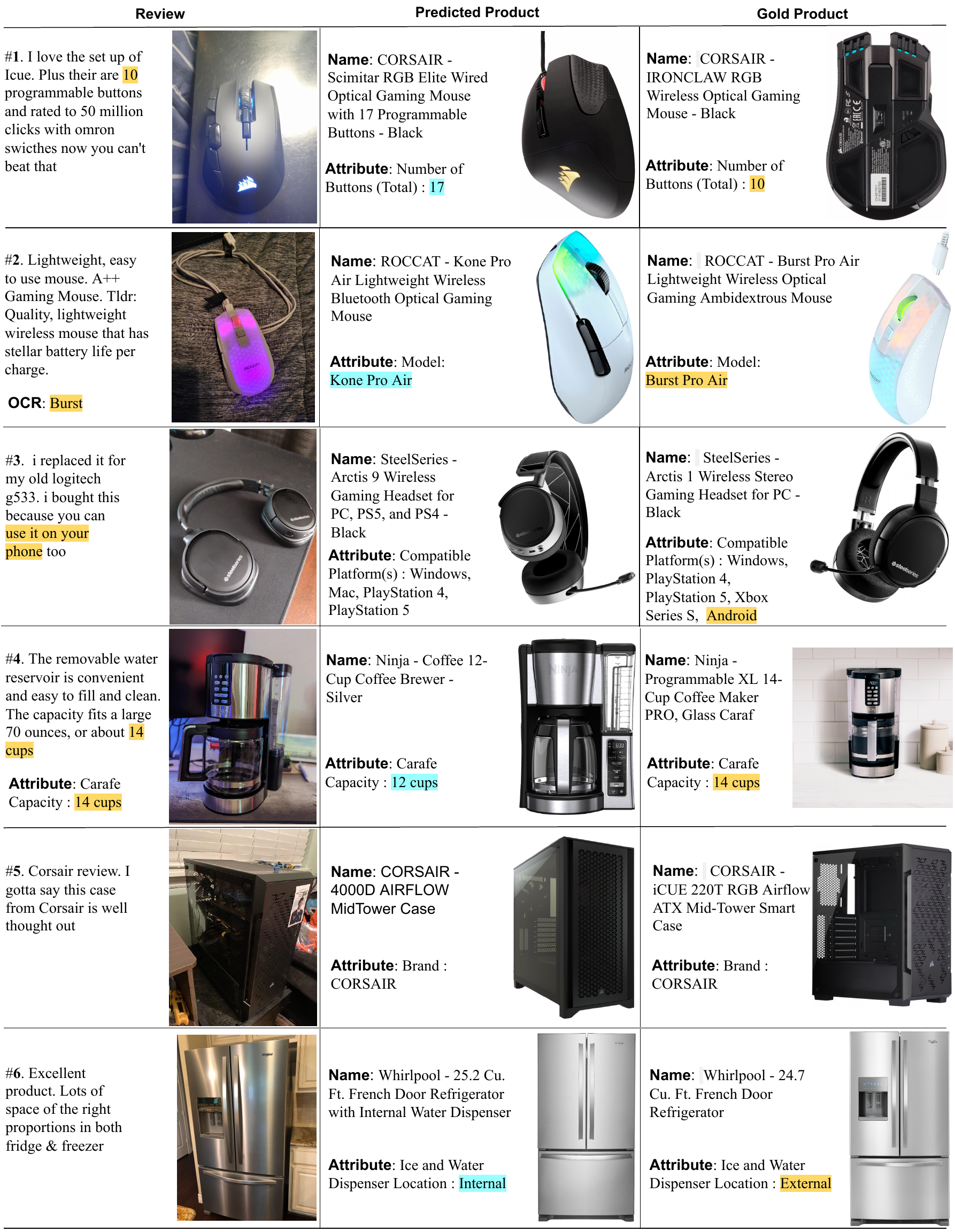}
    \caption{
    Examples of Attribute-aware Multimodal Entity Linking. 
    }
    \label{fig:error_example}
\end{figure*}

We show several error examples of Attribute-aware Multimodal Entity Linking in Figure \ref{fig:error_example}.

\section{Attribute Extraction Performance}\label{sec:attribute_extraction}
\begin{table*}[!ht] 

\small
\centering
\resizebox{1.2\columnwidth}{!}{%
\begin{tabular}{l|c|c|c  }
\toprule
\textbf{ Attribute Extractor} &\textbf{ Precision (\%)} & \textbf{ Recall (\%)} & \textbf{ F1 (\%)}     \\
 \midrule
String Match &97.82 & 37.88& 54.61\\
Zero-shot GPT-2 & 92.38 &37.57 & 53.41 \\
Zero-shot ChatGPT   & 64.57& 17.29& 27.28\\
OCR &98.46  &12.54 &22.24 \\
\midrule
Match+GPT2+ChatGPT+OCR &94.33 &64.18 &76.39 \\
\midrule
Few-shot GPT-2 & 90.04&44.39 & 59.47 \\
Few-shot Vicuna & 74.05& 48.04& 58.28 \\ 
\bottomrule
\end{tabular}%
}
\caption{Performance of attribute value extraction. The term "Match+GPT2+ChatGPT+OCR" signifies the combination of the String Match, zero-shot GPT-2, ChatGPT, and OCR extractor. Due to the computation cost, ChatGPT is only applied for a subset of attribute categories and Vicuna’s max token length is set to 64} 
\label{tab:attribute_extraction}
\vspace{-2mm}
\end{table*}

We have conducted experiments to analyze the performance of each individual attribute extractor and obtained 54.61\%, 53.41\%, 27.28\%, and 22.24\% F-scores on attribute value extraction, corresponding to \textit{String Match}, \textit{zero-shot GPT-2}, \textit{ChatGPT}, and \textit{OCR}, as shown in Table ~\ref{tab:attribute_extraction}. Combining these four extractors leads to a significantly higher F1 score of 76.39\%.  
To shed light on future research, we further conduct experiments of applying \textit{GPT-2}, and an open-source LLM, \textit{Vicuna}~\cite{vicuna2023}, for few-shot attribute extraction, and obtained attribute extraction F1 scores of 59.47\% and 58.28\% on the Test set, respectively. Due to the computation cost, we set Vicuna’s max token length to 64, which may hurt the performance. In this study, we concentrate on establishing the baseline performance for our attribute-aware multimodal entity linking task, and we encourage subsequent research to investigate more advanced methods for extracting and utilizing attribute information.


\section{Application Scenarios} 
\begin{table*}[!ht] 

\small
\centering
\resizebox{2\columnwidth}{!}{%
\begin{tabular}{l|c|c|c |c}
\toprule
\textbf{ Dataset} &\textbf{ Attribute Extractor} & \textbf{ \#Attribute} & \textbf{ \#Mention Context} & \textbf{ Attribute/Mention Ratio} \\
 \midrule
Ours & System & 27533& 16735 & 1.65 \\
Ours - Test Set & Human & 9716& 2741 & 3.54\\
MELBench-Richpedia~\cite{Weibomel}   & System&36705&17800&2.06\\
~\cite{sun2017research} & System* &2198&1000& 2.20\\
~\cite{hu2004mining}  & System* & 348& 314& 1.11\\
\bottomrule
\end{tabular}%
}
\caption{Statistics of attributes within mention context in several datasets. The term "System*" signifies that the attributes have been extracted and documented in the respective work, rather than by our system.} 
\label{tab:dataset_attribute}
\vspace{-2mm}
\end{table*}

To demonstrate the broad application scenario of our proposed attribute-aware entity linking task and approach, we employ both the \textit{String Match} attribute extractor and \textit{Vicuna} attribute extractor on the popularly used entity linking dataset. As illustrated in Table ~\ref{tab:dataset_attribute}, on the Richpedia (MELBench)~\cite{Weibomel} dataset, a public benchmark dataset for multimodal entity linking, the average number of attributes extracted from each mention context is 2.06. Note that the number is only based on the textual descriptions while the images in our multimodal entity linking task may contain more visual attributes. In addition, two other studies~\cite{sun2017research,hu2004mining} have also reported the extraction of 2.20 and 1.11 attributes, respectively, from each mention context within their datasets. Finally, within our dataset, our attribute extractors reveal an average of 1.65 attributes per review, while human annotation yields an average of 3.54 attributes per review. This discrepancy highlights the potential for uncovering more attributes with advanced attribute extractors. Based on these statistics, we respectfully assert that within the Entity Linking (EL) scenario, entity attributes are frequently either explicitly mentioned or implicitly implied within the mention context, and thus, our proposed attribute-aware entity linking task and approach have broad application scenarios.

\section{Experiment Details}
One training for the candidate retrieval model can be done with 1 NVIDIA A40 for 10 hours. One training for the entity disambiguation model can be done with 4 NVIDIA A40 for 7 hours. 
 The search space of hyperparameters for the entity disambiguation model is as follows: the learning rage $\in\{10^{-2}, 10^{-3}, 10^{-4},10^{-5},5\times10^{-3},5\times10^{-4}\}$ and batch size $\in\{12, 16, 20,24,32\}$.

\section{Data Statement}
\subsection{Licensing}
Our dataset is licensed under the CC BY 4.0\footnote{\url{https://creativecommons.org/licenses/by/4.0/}}. The associated codes to \dataset{} for data crawler and baseline are licensed under Apache License 2.0\footnote{\url{https://www.apache.org/licenses/LICENSE-2.0}}.

\subsection{Intended Use}
Our dataset contains products and user reviews in English from E-commerce domains.

The dataset can be used for attribute-aware multimodal entity linking task. A model trained on this task can also be used to link user posts to some products or general entities, a.k.a. detecting user interests from social media.

The dataset can also be used in the unimodal setting, like text-only entity linking.

\subsection{Dataset Format}
Our dataset encompasses a new multimodal entity linking benchmark dataset that contains 16,735 mentions described in text and associated with 30,472 images and a multimodal knowledge base that covers 34,690 entities along with 177,873 entity images and 798,216 attributes.

\begin{enumerate}
\item Multimodal knowledge base
    \begin{enumerate}
        \item Image folder ``product\_images'', which contains all entity images.
        \item Entity information JSON file named ``bestbuy\_products.json'', which contains entity text, image name, and attributes. 
        \begin{enumerate}
            \item product\_category: Category of the product, e.g., ``Video Games -> Nintendo Switch -> Nintendo Switch Games''
            \item product\_name: Name for the product
            \item overview\_section:
                \begin{enumerate}
                    \item description: Description of the product
                \end{enumerate}
            \item image\_path: filename of the corresponding image
            \item image\_url: The link to the corresponding BestBuy image
            \item Spec: Attribute category and attribute value pairs for the product
            \item id: Unique ID for the product
            \item url: The link to the corresponding BestBuy webpage
        \end{enumerate}
    \end{enumerate}
\item Multimodal entity linking dataset, which is split into \textit{Train}, \textit{Dev}, \textit{Test} subsets.
    \begin{enumerate}
        \item Image folder ``review\_images'', which contains all review images.
        \item Image folder ``cleaned\_review\_images''. As explained in Section \ref{sec:preprocess}, review images can also contain irrelevant objects or information. So we apply the object detection model to detect the corresponding object and save the detected image patch as the cleaned review image.
        \item Review information JSON file named ``bestbuy\_reviews.json'', which contains review text, review image name and review attributes. 
        \begin{enumerate}
            \item header: Each review text contains one header and one body
            \item body: Each review text contains one header and one body
            \item mention: The entity mention shown in the review
            \item review\_image\_path: filename of the corresponding review image
            \item review\_image\_url: The link to the corresponding BestBuy image
            \item predicted\_attribute: Review attributes predicted by our attribute extractors
            \item gold\_attribute: Annotated review attributes for the \textit{Test} Set. For the \textit{Train} and \textit{Dev} sets, we clean the predicted\_attribute to obtain gold\_attribute by removing the attributes that do not match with the attributes of the gold entity product.
            \item review\_id: Unique ID for the review
            \item fused\_candidate\_list: Entity IDs for Top-10 candidate entities
            \item gold\_entity\_info
                \begin{enumerate}
                    \item id: Entity ID for the gold entity
                    \item product\_name: Entity name for the gold entity
                    \item product\_category: Entity category for the gold entity
                \end{enumerate}
        \end{enumerate}
    \end{enumerate}
\end{enumerate}




 


\end{document}